\PassOptionsToPackage{table}{xcolor}
\documentclass[lettersize,journal]{IEEEtran}
\usepackage{amsmath,amsfonts}
\usepackage{algorithmic}
\usepackage{algorithm}
\usepackage{array}

\usepackage[caption=false,font=footnotesize,labelfont=rm,textfont=rm]{subfig}

\usepackage{textcomp}
\usepackage{stfloats}
\usepackage{url}
\usepackage{verbatim}
\usepackage{graphicx}
\usepackage{cite}

\usepackage{enumitem}
\usepackage{hyperref}
\usepackage{amsmath}
\usepackage{amssymb}
\usepackage{bm}
\usepackage{threeparttable}
\usepackage{gensymb}
\usepackage{bbding}
\usepackage{makecell}
\usepackage{svg} 
\usepackage{booktabs}
\usepackage{float}
\usepackage{indentfirst}
\usepackage{multirow}
\usepackage{multicol}
\usepackage[table]{xcolor}

\begin{document}

\title{
LI-GS: Gaussian Splatting with LiDAR Incorporated for Accurate Large-Scale Reconstruction
}

\author{
	Changjian Jiang$^{1}$,
	Ruilan Gao$^{1}$,
        Kele Shao$^{1}$,
        Yue Wang$^{1}$,
        Rong Xiong$^{1}$,
	Yu Zhang$^{1,2,\ast}$ 

    \thanks{The project page is: \url{https://changjianjiang01.github.io/LI-GS/} .}
    \thanks{This research was supported by National Key R\&D Program of China under Grant 2023YFB4704400, in part by Zhejiang Provincial Natural Science Foundation of China under Grant No. LD24F030001, and in part by NSFC 62088101 Autonomous Intelligent Unmanned Systems.}
    \thanks{$^{1}$ State Key Laboratory of Industrial Control Technology, College of Control Science and Engineering, Zhejiang University, Hangzhou, China.}
    \thanks{$^{2}$ Key Laboratory of Collaborative sensing and autonomous unmanned systems of Zhejiang Province, Hangzhou, China.}	
    \thanks{$^{\ast}$ Correspondence: Yu Zhang. Email: zhangyu80@zju.edu.cn}
    }


\IEEEpubid{0000--0000/00\$00.00~\copyright~2021 IEEE}

\maketitle

\begin{abstract}
Large-scale 3D reconstruction is critical in the field of robotics, and the potential of 3D Gaussian Splatting (3DGS) for achieving accurate object-level reconstruction has been demonstrated. However, ensuring geometric accuracy in outdoor and unbounded scenes remains a significant challenge.
This study introduces LI-GS, a reconstruction system that incorporates LiDAR and Gaussian Splatting to enhance geometric accuracy in large-scale scenes. 2D Gaussain surfels are employed as the map representation to enhance surface alignment. Additionally, a novel modeling method is proposed to convert LiDAR point clouds to plane-constrained multimodal Gaussian Mixture Models (GMMs). The GMMs are utilized during both initialization and optimization stages to ensure sufficient and continuous supervision over the entire scene while mitigating the risk of over-fitting.
Furthermore, GMMs are employed in mesh extraction to eliminate artifacts and improve the overall geometric quality.
Experiments demonstrate that our method outperforms state-of-the-art methods in large-scale 3D reconstruction, achieving higher accuracy compared to both LiDAR-based methods and Gaussian-based methods with improvements of 52.6\% and 68.7\%, respectively.
\end{abstract}


\section{Introduction}
Navigation in large-scale outdoor environments is crucial in various robotic applications, such as autonomous driving \cite{cui2021deep}, industrial inspection \cite{tao2024silvr}, and embodied AI \cite{wang2024embodiedscan}. 
The reconstruction of an accurate and dense 3D map of the environment plays an essential role to achieve these tasks.
LiDAR-visual fusion is a prominent technique for large-scale 3D reconstruction that combines the geometric information provided by LiDAR with the texture information captured by cameras.

\begin{figure}[!t]
    \centering
    \includegraphics[width=0.93\linewidth]{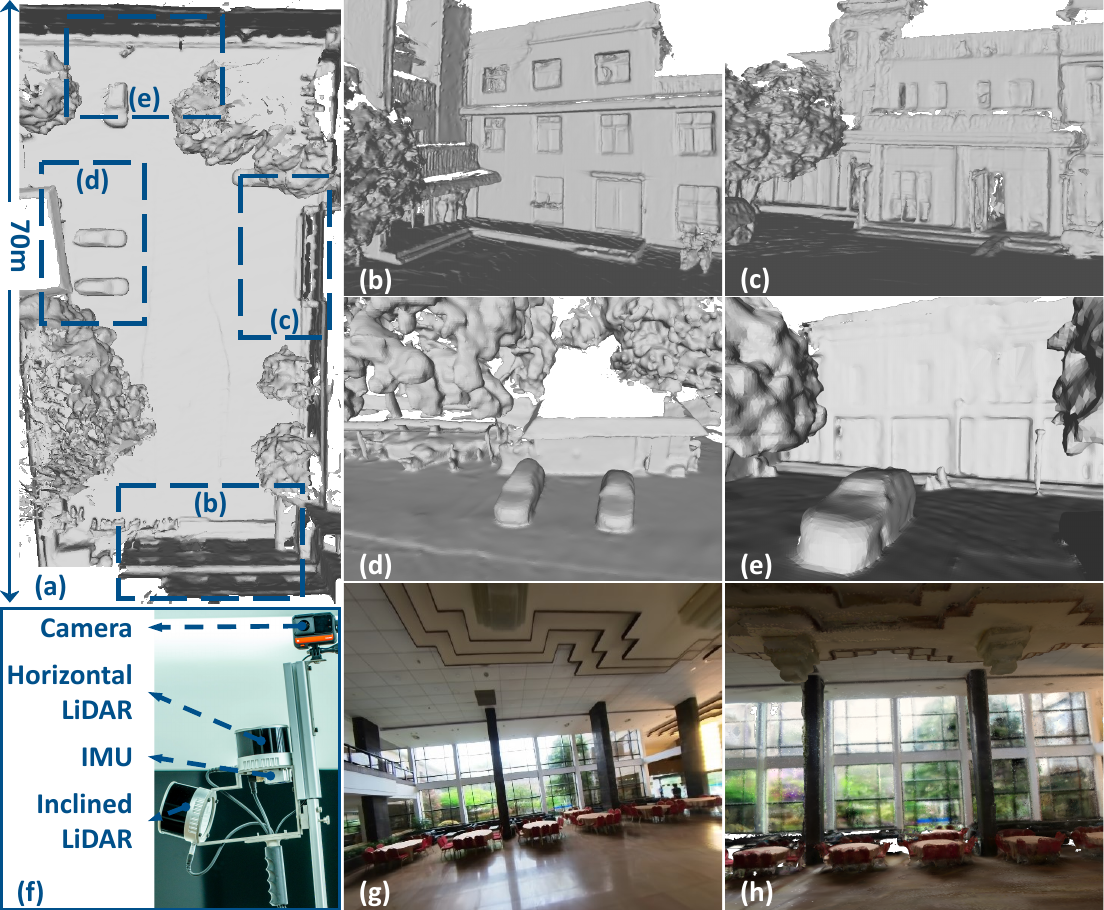}
    \caption{The performance of LI-GS. (a) The comprehensive mesh of a campus scene, with its details shown in (b)-(e). (f) Our data collection platform. (g)-(h) The rendered RGB image and the colorized mesh of an indoor scene.}
    \label{fig:first_image}
\end{figure}

Recently, 3D Gaussian Splatting (3DGS) \cite{kerbl20233d} has gained widespread attention in 3D reconstruction. In contrast to neural implicit representations such as neural radiance field (NeRF) 
\cite{mildenhall2021nerf,isaacson2023loner,rematas2022urban} and neural signed distance function (SDF) 
\cite{zhong2023shine,deng2023nerf,yu2023nf}, 3DGS explicitly represents the environment, leading to significant reduction in training time and enabling real-time rendering. These advantages position 3DGS as a promising map representation, as shown in studies \cite{keetha2024splatam, matsuki2024gaussian,wu2024mm}.
However, 3DGS encounters challenges in achieving geometrically accurate reconstruction, especially in sparse-view, unbounded and large-scale scenes. These challenges can be attributed to three main factors: 
\textbf{(1) Ellipsoid-like shape.} Gaussians exhibit ellipsoid-like shapes, which violates the assumption of thin object surfaces, resulting in poor surface fitting.  
\textbf{(2) Lack of precise depth information.} The lack of precise depth information hinders the supervision along the camera's principal axis. As a result, the photometric loss excessively impacts the geometric attributes of the Gaussians, leading to their inaccurate placement.
\textbf{(3) Sparse supervise views.} Autonomous systems often capture a limited number of supervise views, and exclusively performing object-centric trajectories for improving reconstruction is infeasible. Consequently, 3DGS tends to overfit to a single view and lacks geometric consistency across multiple views.

\IEEEpubidadjcol
Current studies \cite{huang20242d, dai2024high} directly set the z-scale of Gaussians to zero, which flattens 3D ellipsoids into 2D surfels, facilitating object-level reconstruction. However, the lack of accurate depth information hinders these methods from attaining desirable results in large-scale scenes.
Certain studies \cite{hong2024liv,zhou2024hugs,cui2024letsgo} integrate LiDAR with 3DGS. Nonetheless, the sparsity of supervise views limits LiDAR's capacity to provide comprehensive constraints for the entire scene, leading to inferior mapping effects.
In summary, reconstructing accurate surfaces using Gaussians in large-scale scenes remains a challenge.

\begin{figure*}[!t]
    \centering
    \includegraphics[width=0.95\linewidth]{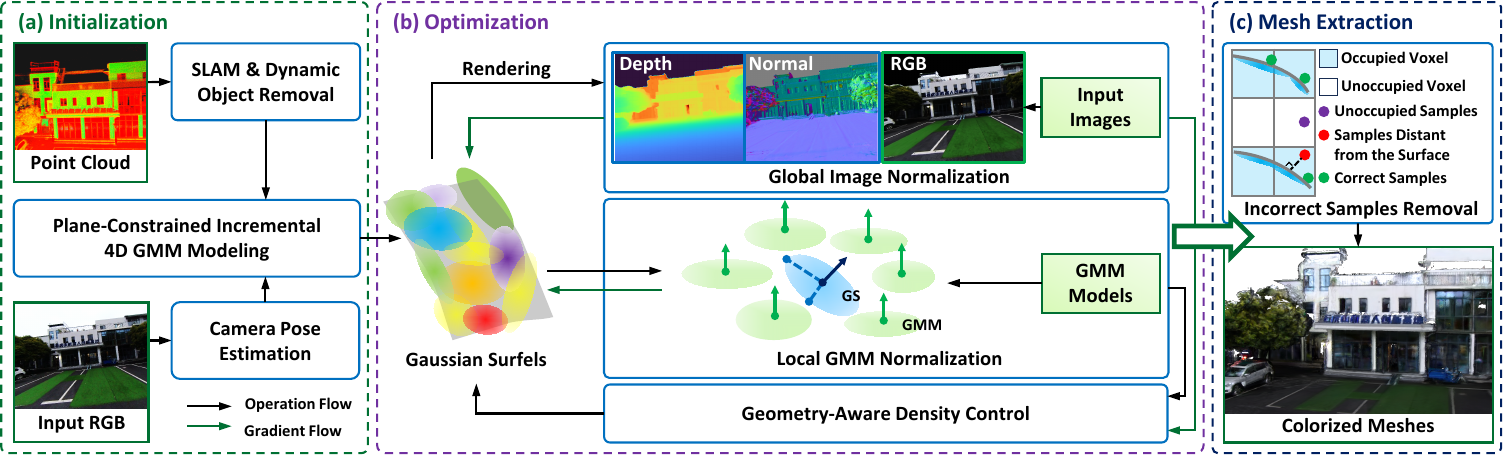}
    \caption{The overview of our system. Our system involves three steps: (a) We utilize the plane-constrained incremental 4D GMM to generate initial Gaussian surfels. (b) To optimize the Gaussian surfels, we apply global image normalization, local GMM normalization, and geometry-aware density control. (c) We eliminate incorrect samples using a coarse-to-fine method, and then apply the screened Poisson reconstruction method \cite{kazhdan2013screened} to extract meshes.}
    \label{fig:system}
\end{figure*}

To tackle this challenge, we propose LI-GS, a reconstruction system that incorporates LiDAR with Gaussian Splatting to enhance geometric accuracy in large-scale scenes, as shown in Fig. \ref{fig:first_image}. 
Our system utilizes 2D Gaussain surfels \cite{huang20242d} as the map representation to improve surface alignment. 
We consider both LiDAR point clouds and Gaussian surfels as samples drawn from a probability distribution that represents the surfaces of scenes. 
LiDAR point clouds, serving as an initial sparse sampling, provide precise geometric information to guide the resampling of Gaussian surfels, resulting in accurate reconstruction. 

Specifically, plane-constrained multimodal Gaussian Mixture Models (GMMs) are utilized to convert LiDAR point clouds to a continuous probabilistic model. The utilization of these GMMs offers several advantages. 
Firstly, plane-constrained multimodal GMMs can serve as initial models that accurately capture the geometry and textures of scenes, which are not attainable through SfM \cite{schonberger2016structure}. 
Secondly, GMMs can enhance constraints along the camera's principal axis during optimization and eliminate artifacts near depth discontinuities during mesh extraction. 
Thirdly, GMMs provide continuous and sufficient supervision in unobserved areas and prevent over-fitting to a single view.
Our system is comprehensively compared with thirteen excellent methods. The experimental results demonstrate that our method attains state-of-the-art performance in large-scale 3D reconstruction.
Overall, the contributions of this paper can be summarized as follows: 
\begin{itemize}
  \item A LiDAR-visual reconstruction system is proposed, which incorporates LiDAR during both initialization and optimization stages of Gaussian Splatting.
  \item The plane-constrained multimodal GMM is introduced, which converts colorized point clouds into accurate initial Gaussians.
  \item A geometric normalization method is introduced, which leverages GMMs to optimize the attributes and distribution of Gaussians.
  \item The system is evaluated in various scenarios and achieves significant improvements compared to LiDAR-based methods (52.6\%) and Gaussian-based methods (68.7\%).
\end{itemize}

\section{Related Works}
\subsection{Large-Scale 3D Reconstruction}
In contrast to the explicit 3D representations shown in \cite{ruan2023slamesh, vizzo2022vdbfusion}, several studies have embraced neural implicit representations for LiDAR-based outdoor 3D reconstruction \cite{isaacson2023loner, zhong2023shine, deng2023nerf, yu2023nf}. 
These approaches employ multilayer perceptrons (MLPs) to encapsulate the entire scene and achieve satisfactory results. Nonetheless, the sparsity of LiDAR scans may compromise the reconstruction quality, especially in areas with limited LiDAR coverage.
In addition, some implicit methods adopt a LiDAR-visual fusion approach for large-scale scene reconstruction. Urban Radiance Fields \cite{rematas2022urban} presents a method that utilizes LiDAR to extend the application of the NeRF to large-scale scenes. Inspired by this approach, SiVLR \cite{tao2024silvr} integrates with a SLAM system and employs RGB images, LiDAR depth, and normal images for training a NeRF.
In contrast to neural implicit representations, 3D Gaussian Splatting (3DGS) \cite{kerbl20233d} utilizes a collection of Gaussians with learnable attributes to explicitly represent appearance and geometry, thereby enabling real-time rendering.

\subsection{Geometrically Accurate Reconstruction Using Gaussians}
Numerous Gaussian-based methods have been proposed to enhance the geometric accuracy of 3D reconstruction. 
2DGS \cite{huang20242d} and Gaussian Surfels \cite{dai2024high} propose to flatten Gaussian ellipsoids to surfels and introduce a normal-depth consistency regularization. Additionally, SuGaR \cite{guedon2024sugar} employs the SDF to enforce the alignment of Gaussians with object surfaces. 
PGSR \cite{chen2024pgsr} renders unbiased depth maps and proposes a multi-view regularization. In addition to solely focusing on regularization, Trim3DGS \cite{fan2024trim} propose a contribution-based trimming strategy to eliminate inaccurate Gaussians.
GOF \cite{yu2024gaussian} employs ray-tracing-based volume rendering to enable the direct extraction of geometry. Building upon GOF \cite{yu2024gaussian}, RaDe-GS \cite{zhang2024rade} introduces a rasterized approach for computing precise depth maps of general Gaussian splats. 
Furthermore, there are methods that combine 3DGS with neural SDF \cite{yu2024gsdf, chen2023neusg, lyu20243dgsr}.
Regarding mesh extraction, Gaussian Surfels involve constructing an occupancy map using Gaussians and removing sampled points within unoccupied voxels. 

However, the aforementioned methods face challenges in dealing with large-scale scenes, even when LiDAR is integrated for initialization. Therefore, we propose a novel GMM-based normalization approach to accurately learn geometry. Additionally, we present a coarse-to-fine method for efficient mesh extraction.

% ER-mapping effect
\begin{figure}[!t]
    \centering
    \includegraphics[width=0.88\linewidth]{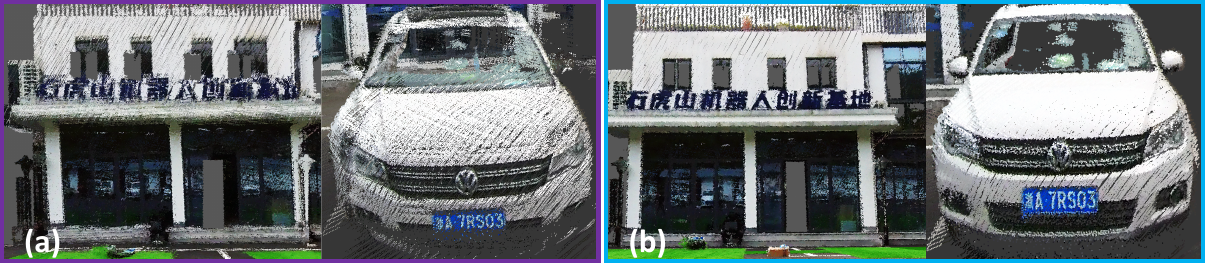}
    \caption{Comparison of colorized point clouds generated using two different methods, (a) interpolated image poses and (b) our previous work \cite{jiang2024er}. Our method produces colorized point clouds with more distinct textures.
    }
    \label{fig:camera_pose}
\end{figure}

\section{Preliminaries}
2DGS \cite{huang20242d} models the scene using a set of flat Gaussian surfels, allowing for better alignment with thin surfaces. The basic attributes of a surfel include the central point \(\mathbf{p}_i \in \mathbb{R}^{3}\), radii \(r_{u_i} \ge r_{v_i} \in \mathbb{R}^{+}\), two corresponding normalized orthogonal vectors \(\mathbf{t}_{u_i}\), \(\mathbf{t}_{v_i}\in\mathbb{R}^{3}\), a normal vector  \(\mathbf{n}_{i}=\mathbf{t}_{u_i}\times\mathbf{t}_{v_i}\), opacity \(o_i\in[0,1]\), and the view-dependent appearance \(\mathbf{c}_i\in\mathbb{R}^3\) parameterized with spherical harmonics.
A Gaussian surfel defines a local 2D tangent space. A point \(\mathbf{u}=[u,v]^{\top}\) in this space can be converted into the world space \(\mathbf{p}(\mathbf{u})\in \mathbb{R}^{3}\) via 
$\mathbf{p}(\mathbf{u})=\mathbf{p}_i + r_{u_i}\mathbf{t}_{u_i}u + r_{v_i}\mathbf{t}_{v_i}v\text{.}$
Its Gaussian value \(f(\mathbf{u})\) is evaluated as $f(\mathbf{u})=\exp\left(-{(u^2+v^2)} \big / {2}\right)$.

To render an image, 2DGS first sorts the surfels in a front-to-back order. For a point \(\mathbf{x} \in \mathbb{R}^{2}\) in the image space, its appearance \(\mathbf{c}(\mathbf{x})\) is calculated via
\begin{equation}
    \mathbf{c}(\mathbf{x})=\sum_{i=1}^{N}\mathbf{c}_i o_i f(\mathbf{u}_i(\mathbf{x}))\prod_{j=1}^{i-1}(1- o_j f(\mathbf{u}_j(\mathbf{x})))\text{,}
\end{equation}
where \(N\) is the number of visible surfels, and \(\mathbf{u}_i(\mathbf{x})\) is a 2D-to-2D mapping that projects a point in the image space onto the tangent space by finding the intersection of three non-parallel planes.

\section{Methodology}
Fig. \ref{fig:system} provides an overview of our system. In this section, we introduce our system from the following aspects: preprocessing (\ref{subsection:Preprocessing}), initialization (\ref{subsection:Init}), optimization (\ref{subsection:Optimization}), and mesh extraction (\ref{subsection:Meshing}).

\subsection{Preprocessing}
\label{subsection:Preprocessing}
Our system processes LiDAR scans and RGB images as inputs and incorporates preprocessing steps to enhance data accuracy. Initially, SLICT \cite{nguyen2023slict}, a state-of-the-art LiDAR-Inertial continuous-time SLAM system, is employed to estimate the initial poses of the LiDAR scans. 
Subsequently, M-detector \cite{wu2024moving} is utilized to remove dynamic objects from the scans, followed by HBA \cite{liu2023large} to enhance the accuracy of the global point cloud.
For input images, our previous work \cite{jiang2024er} provides accurate initial image poses, as shown in Fig. \ref{fig:camera_pose}. These poses are further refined using Colmap-PCD \cite{bai2024colmap}. 
The preprocessing steps result in accurate global point clouds and consistent image poses.

% ISOGMM pipeline
\begin{figure}[!t]
    \centering
    \includegraphics[width=0.86\linewidth]{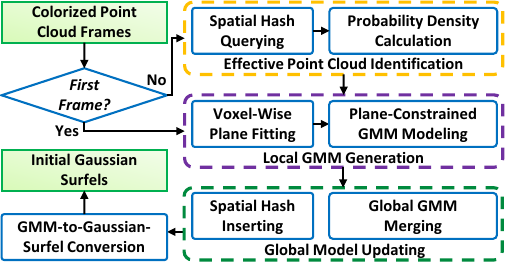}
    \caption{The pipeline of plane-constrained incremental 4D GMM modeling. 
    }
    \label{fig:gmm_system}
\end{figure}

\subsection{Initialization}
\label{subsection:Init}
Building upon previous works \cite{goel2023probabilistic,goel2023incremental}, we propose a method for generating multimodal Gaussian Mixture Models (GMMs) from large-scale colorized point clouds and utilize the spatial hash to efficiently maintain the GMMs. The pipeline is shown in Fig. \ref{fig:gmm_system}.

First, we traverse the images and project the global point cloud, generating a sequence of colorized point cloud frames.
For the first frame, voxelization is performed followed by the utilization of RANSAC \cite{fischler1981random} within each voxel to extract planes. 
A set of points located on the same plane can be denoted as \(\mathcal{P}=\{\mathbf{z}_i \mid \mathbf{z}_i=[\mathbf{p}_i,g_i]^{\top},\mathbf{p}_i\in\mathbb{R}^{3}, g_i\in[0,1] \}\), where \(\mathbf{p}_i\) signifies the position in the world space and \(g_i\) represents the gray scale calculated using the RGB value. 
These points can be  characterized by their mean \(\overline{\mathbf{p}}\in\mathbb{R}^{3}\), eigenvalues \(\alpha_0 \le \alpha_1 \le \alpha_2\) and corresponding normalized eigenvectors \(\mathbf{v}_0\), \(\mathbf{v}_1\), and \(\mathbf{v}_2\) of their covariance.

\begin{figure}[!t]
    \centering
    \includegraphics[width=0.85\linewidth]{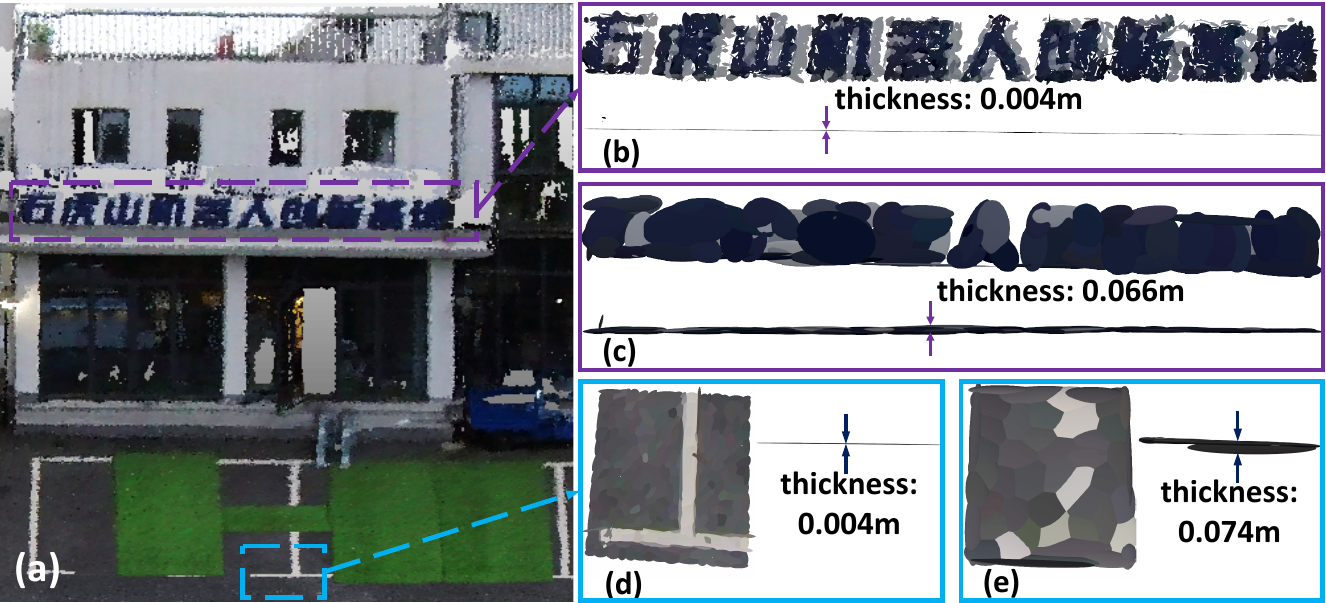}
    \caption{The performance of plane-constrained 4D GMM modeling. GMM components are represented as ellipsoids. (a) Illustration of the input colorized point cloud. (b), (d) The plane-constrained 4D GMMs in two specific areas. (c), (e) The 3D GMMs without plane constraints.}
    \label{fig:4d_plane}
\end{figure}

The local 4D GMM is modeled from the point set \(\mathcal{P}^{\prime}=\{\mathbf{z}^{\prime}_i \mid \mathbf{z}^{\prime}_i=[u_i,v_i,0,g_i]^{\top} \}\) in the plane frame, where \(\mathbf{p}_i = \overline{\mathbf{p}}+u_i\mathbf{v}_2+v_i\mathbf{v}_1+w_i\mathbf{v}_0\). The probability density of a point \(\mathbf{z}^{\prime}=[u,v,0,g]^{\top}\) located on this plane can be calculated via
\begin{equation}
    p_L(\mathbf{z}^{\prime}) = \sum_{l\in\mathcal{L}}\pi^{\prime}_l\mathcal{N}(\mathbf{z}^{\prime} \mid \bm{\mu}^{\prime}_l,\bm{\Sigma}^{\prime}_l)\text{,}
\end{equation}
where \(\pi^{\prime}_l\), \(\bm{\mu}^{\prime}_l\), and \(\bm{\Sigma}^{\prime}_l\in\mathbb{S}^{4\times4}\) are the weight, mean, and covariance of the GMM component \(l\), respectively. The corresponding set of indices is denoted as \(\mathcal{L} \). 
Notably, \(\pi^{\prime}_l\) satisfies \(\sum_{j\in\mathcal{L}}\pi^{\prime}_l = 1\).
Moreover, a component can be transformed into the world space via
\begin{equation}
    \begin{aligned}
           \pi_l = \pi^{\prime}_l\text{, }\bm{\mu}_l = [\overline{\mathbf{p}}^\top,0]^\top + \mathbf{H}\bm{\mu}^{\prime}_l\text{, }\bm{\Sigma}_l = \mathbf{H}\bm{\Sigma}^{\prime}_l\mathbf{H}^{\top} \text{, }\\
           \mathbf{H}=\begin{bmatrix}\mathbf{R}&&\mathbf{0}\\\mathbf{0}^{\top}&&1\end{bmatrix}\text{, } \mathbf{R}=[\mathbf{v}_2, \mathbf{v}_1,\mathbf{v}_0]\in\rm{SO}(3)\text{, }   
    \end{aligned}
\end{equation}
where \(\pi_l\), \(\bm{\mu}_l\), and \(\bm{\Sigma}_l\in\mathbb{S}^{4\times4}\) are the weight, mean, and covariance in the world space. As shown in Fig. \ref{fig:4d_plane}, 4D GMMs take into account the distribution of points in the color dimension, accurately representing the surface textures. Additionally, plane constraints facilitate effective noise removal, leading to thinner components that are highly suitable for conversion into Gaussian surfels.

The number of GMM components, denoted by \(\left|\mathcal{L}\right|\), is adjusted based on the scene complexity. It is determined by minimizing an information-theoretic objective function as described in \cite{goel2023probabilistic}. This objective function can be approximated using the Gaussian Mean Shift \cite{cheng1995mean}.
 
Upon receiving a subsequent colorized point cloud frame \(\mathcal{F}\), the system identifies the effective points in the new space.
The points in \(\mathcal{F}\) are divided into two sets: the ones in the existing voxels (\(\mathcal{F}^\text{o}\)) and the ones in the new voxels (\(\mathcal{F}^\text{n}\)) by querying the spatial hash. They satisfy the condition \(\mathcal{F} = \mathcal{F}^\text{n}\cup\mathcal{F}^\text{o}\).
% log-likelihood
The effective points in the existing voxels (\(\mathcal{F}^\text{l}\)) are determined by calculating the log-likelihood as
\begin{equation}
    \mathcal{F}^{\text{l}} = \{\mathbf{z}_j\in\mathcal{F}^{\text{o}}\mid\mathbf{z}_j=[\mathbf{p}_j,g_j]^{\top},L(\mathbf{p}_j)<\rho\}\text{,}  
\end{equation} 
\begin{equation}
\begin{aligned}
    L(\mathbf{p}_j) &= \ln\left(p_K\left(\mathbf{p}_j\right) \right) \\
                    &= \ln\left(\sum_{k\in\mathcal{K}}\pi_k\mathcal{N}(\mathbf{p}_j\mid\bm{\mu}_{k}^{\mathbf{p}},\bm{\Sigma}_k^\mathbf{pp})\right), \\
\end{aligned}
\end{equation} 
\begin{equation}
    \bm{\mu}_{k} = [\bm{\mu}_{k}^{\mathbf{p}},\bm{\mu}_{k}^{g}]^{\top}\text{, }\bm{\Sigma}_{k}=
    \begin{bmatrix}
        \bm{\Sigma}_{k}^{\mathbf{pp}} & \bm{\Sigma}_{k}^{\mathbf{p}g} \\ 
        \bm{\Sigma}_{k}^{g\mathbf{p}} & \bm{\Sigma}_{k}^{gg}
    \end{bmatrix}
    \text{,} 
\end{equation}
where \(\mathcal{K}\) denotes the set of indices of the global GMM. \(\rho\) is a pre-determined threshold, and \(L(\mathbf{p}_j)\) denotes the log-likelihood of \(\mathbf{p}_j\). The distribution of gray scale is unrelated to spatial effectiveness, so \(L(\mathbf{p}_j)\) is calculated using the marginal probability density \(p_K\left(\mathbf{p}_j\right)\) instead of \(p_K\left(\mathbf{z}_j\right)\). Consequently, the effective points in the current frame (\(\mathcal{F}^\text{e}\)) are determined as \(\mathcal{F}^\text{e} = \mathcal{F}^\text{n}\cup\mathcal{F}^\text{l}\). 
After traversing all the images, the global GMM is converted to the initial Gaussian surfels via
\begin{equation}
    \begin{aligned}
      &\mathbf{p}_{k} =\bm{\mu}_{k}^{\mathbf{p}} \text{, }  
      \mathbf{n}_{k} = \mathbf{w}_{0_k} \text{, }
      \mathbf{t}_{u_k} = \mathbf{w}_{2_k} \text{, }
      \mathbf{t}_{v_k} = \mathbf{w}_{1_k}\text{, }\\ 
      &r_{u_k} = \sqrt{\gamma_{2_k}} \text{, }
      r_{v_k} = \sqrt{\gamma_{1_k}} \text{, }
      o_{k}=0.6 + 0.4\pi_k\text{, }
    \end{aligned}
\end{equation}
where \(\gamma_{0_k} \le \gamma_{1_k} \le \gamma_{2_k}\) and \(\mathbf{w}_{0_k}\), \(\mathbf{w}_{1_k}\), and \(\mathbf{w}_{2_k}\) are the eigenvalues and corresponding eigenvectors of \(\bm{\Sigma}_{k}^{\mathbf{pp}}\), respectively.

\subsection{Optimization}
\label{subsection:Optimization}

Despite the introduction of LiDAR in initialization, both 3DGS \cite{kerbl20233d} and 2DGS \cite{huang20242d} produce noisy reconstruction when solely optimized with photometric loss, especially in sparse-view scenes. To address this issue, we propose a comprehensive normalization approach alongside an innovative geometry-aware density control method.

\subsubsection{Normalization}
Our total loss $L$ consists of five components: GMM loss $L_\text{GMM}$, photometric loss \(L_\text{p}\), sky loss \(L_\text{sky}\), depth image loss \(L_\text{d}\), and normal image
loss \(L_\text{n}\), which can be calculated via 
\begin{equation}
    L = \lambda_\text{GMM}L_\text{GMM} + L_\text{p} + L_\text{sky} + \lambda_\text{d}L_\text{d} + \lambda_\text{n}L_\text{n}\text{,}
\end{equation}
where weights \(\lambda_\text{GMM}\), \(\lambda_\text{d}\), and \(\lambda_\text{n}\) balance the loss terms.

\begin{figure}[!t]
    \centering
    \includegraphics[width=0.85\linewidth]{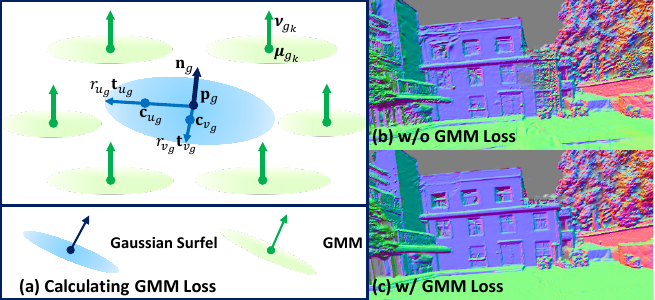}
    \caption{Using GMMs to supervise Gaussian surfels. (b)-(c) Comparison of rendered normal images without and with GMM loss. The GMM loss effectively mitigates the noise present in the result surfaces.
    }
    \label{fig:gmm_gs}
\end{figure}

\textbf{GMM loss.} The multimodal GMM is employed to optimize the position and shape of Gaussian surfels in 3D space.
As shown in Fig. \ref{fig:gmm_gs}(a), \(\mathbf{p}_g\), \(\mathbf{n}_g\), \(r_{u_g} \ge r_{v_g} \), \(\mathbf{t}_{u_g}\), and \(\mathbf{t}_{v_g}\) represent the position, normal, radii and corresponding principle vectors of the Gaussian surfel indexed with \(g\), respectively. 
Following this, the \(K\) nearest GMM components are identified, where \(\bm{\mu}_{g_k}\) and \(\bm{\nu}_{g_k}\) represent the mean and normal vector of the GMM component \(g_k\).

Firstly, the distance \(L_\text{dis}\) between \(\mathbf{p}_g\) and the surface defined by the \(K\) GMM components is minimized to ensure proper alignment of the Gaussian surfels with the local structure. The \(L_\text{dis}\) is calculated via
\begin{equation}
      L_\text{dis} = \frac{1}{G}\sum_{g=1}^{G}d_g(\mathbf{p}_g)\text{, } 
\end{equation}
where \(G\) represents the number of visible Gaussian surfels under the current viewpoint, and \(d_g(\mathbf{p})\) is a weighted distance 
% from a point to the plane, calculated as:
\begin{equation}
d_g(\mathbf{p}) = \sum_{k=1}^{K}\omega_{g_k}\left\|(\mathbf{p}-\bm{\mu}_{g_k})^\top\bm{\nu}_{g_k}\right\|_1,
\end{equation}
where the weight \(\omega_{g_k} = \exp\left(-{\left\|\mathbf{p}_g-\bm{\mu}_{g_k}\right\|^2_2} \big / {2\sigma^2}\right)\) emphasizes the contribution of closer GMM components.

% L_control
To ensure the shape accuracy of Gaussian surfels, we introduce shape control points \(\mathbf{c}_{u_g}=\mathbf{p}_g+\alpha r_{u_g} \mathbf{t}_{u_g}\) and \(\mathbf{c}_{v_g}=\mathbf{p}_g+\alpha r_{v_g} \mathbf{t}_{v_g}\), and we minimize the distance \(L_\text{control}\) calculated via
\begin{equation}
    L_\text{control} 
    = \frac{1}{G}\sum_{g=1}^{G}l_g \text{, } 
\end{equation}
\begin{equation}
    l_g  = 
    \begin{cases}
        d_g(\mathbf{c}_{u_g})+d_g(\mathbf{c}_{v_g}),&{\text{if}} \ r_{v_g} \ge \phi, \\ 
        d_g(\mathbf{c}_{u_g}),&{\text{if}} \ {r_{u_g} \ge \phi,\ r_{v_g} \le \phi,}     \\
        0,&{\text{otherwise,}}
    \end{cases}
\end{equation}
where the threshold \(\phi\) serves to selectively supervise larger Gaussian surfels.
Additionally, we leverage the weighted normal vectors to supervise the normal vectors of Gaussian surfels through the loss function \(L_\text{normal}\), which is computed via
%derived from the GMMs 
\begin{equation} 
    L_\text{normal}=\frac{1}{G}\sum_{g=1}^{G}\left\|\mathbf{n}_g-\bar{\mathbf{n}}_g\right\|_1+\left\|1-\mathbf{n}_g^\top\bar{\mathbf{n}}_g\right\|_1\text{,} 
\end{equation}
where \(\bar{\mathbf{n}}_g = \left(\sum_{k=1}^{K}\omega_{g_k}\bm{\nu}_{g_k} \right)\Big / \left\|\sum_{k=1}^{K}\omega_{g_k}\bm{\nu}_{g_k}\right\|_2 \). Finally,  \(L_\text{GMM}\) is calculated via \(L_\text{GMM}=L_\text{dis}+L_\text{control}+L_\text{normal}\). 

% Photometric loss
\textbf{Photometric loss.} Similar to 3DGS \cite{kerbl20233d}, we utilize the photometric loss to minimize the difference between the rendered RGB image \(\tilde{\mathbf{I}}\) and the input image \(\mathbf{I}\) via 
\begin{equation}
    L_{\text{p}}=0.8L_1(\tilde{\mathbf{I}},\mathbf{I})+0.2L_{\text{D-SSIM}}(\tilde{\mathbf{I}},\mathbf{I})\text{.}
\end{equation}

\textbf{Sky loss.} 
The sky loss is introduced in outdoor scenes to mitigate artifacts specifically within the sky region. 
Firstly, a semantic segmentation network \cite{mmseg2020} is utilized to generate a sky mask, denoted as \(\mathbf{M}\), where zero indicates sky regions. \(\mathbf{M}\) is employed to mask the sky area in the input RGB image.
Furthermore, the sky loss \(L_{\text{sky}}\) is calculated via 
\begin{equation}
        L_{\text{sky}} =\left(1 - \mathbf{M}\right)L_1(\tilde{\mathbf{S}})\text{,} 
\end{equation}
where \(\tilde{\mathbf{S}}\) is the rendered silhouette image.

\textbf{Depth and normal image losses.} Depth images \(\hat{\mathbf{D}}\) and normal images \(\hat{\mathbf{N}}\) derived using LiDAR are employed to align Gaussian surfels with the global structure via
\begin{equation}
        L_{\text{d}} =L_1(\tilde{\mathbf{D}},\hat{\mathbf{D}})\text{, } 
        L_{\text{n}} =(1-\tilde{\mathbf{N}}\cdot\hat{\mathbf{N}})\text{,} 
\end{equation}
where \(\tilde{\mathbf{D}}\) and \(\tilde{\mathbf{N}}\) denote the rendered depth and normal images, respectively.

\subsubsection{Geometry-Aware Density Control}
Our system dynamically controls both the number and density of Gaussians throughout the optimization stage, enabling a transition from a sparse set of initial Gaussians to a denser set that provides a more accurate representation of the scene.
To eliminate redundant and inaccurate Gaussians while improving the distribution of Gaussians according to the geometric structure, we introduce the weighted distance \(d_g(\mathbf{p}_g)\) into both the growing and pruning mechanisms, enhancing the gradient-based density control criteria used in 3DGS \cite{kerbl20233d}. 
Specifically, the growing criterion is formulated as
\begin{equation}
    \epsilon^\text{growth}_g=(1-\omega_\text{growth})\nabla_g+\omega_\text{growth}\omega_\text{scale}\exp\left(-\frac{d_g(\mathbf{p}_g)^2}{2\tau^2}\right)\text{,}
\end{equation}
where \(\nabla_g\) represents the averaged position gradient, and \(\omega_\text{scale}\) is employed to ensure that the magnitudes are on a similar scale. Consequently, when \(\epsilon^\text{growth}_g\) exceeds a predetermined threshold, a new Gaussian surfel is added.
Furthermore, Gaussian surfels far away from the surfaces are more prone to be pruned. The pruning criterion is defined as 
\begin{equation}
    \epsilon^\text{pruning}_g=o_g-\omega_\text{pruning}\left(1-\exp\left(-\frac{d_g(\mathbf{p}_g)^2}{2\tau^2}\right)\right)\text{,}
\end{equation}
where \(o_g\) denotes opacity. Gaussian surfels exhibiting low \(\epsilon^\text{pruning}_g\) are subsequently pruned.

\subsection{Mesh Extraction}
\label{subsection:Meshing} 
Gaussian Surfels \cite{dai2024high} construct an occupancy map and remove sampled points within unoccupied voxels to eliminate artifacts near depth discontinuities. Nonetheless, in unbounded scenes, large grids may result in excessive point removal, while small grids compromise computational efficiency. To tackle this issue, we incorporate GMMs into mesh extraction.

Initially, we construct a voxel map and compute the occupancy. Samples within unoccupied voxels (purple points in Fig. \ref{fig:system}(c)) are then removed. Subsequently, to facilitate refinement, we discard samples located distant from the object surfaces within occupied grids (red points in Fig. \ref{fig:system}(c)), based on the weighted distance \(d(\mathbf{p})\). Finally, the mesh is extracted using the screened Poisson method \cite{kazhdan2013screened}.

\begin{figure*}[!t]
\centering
\subfloat[LI-GS (Ours)]{\includegraphics[width=0.14\linewidth]{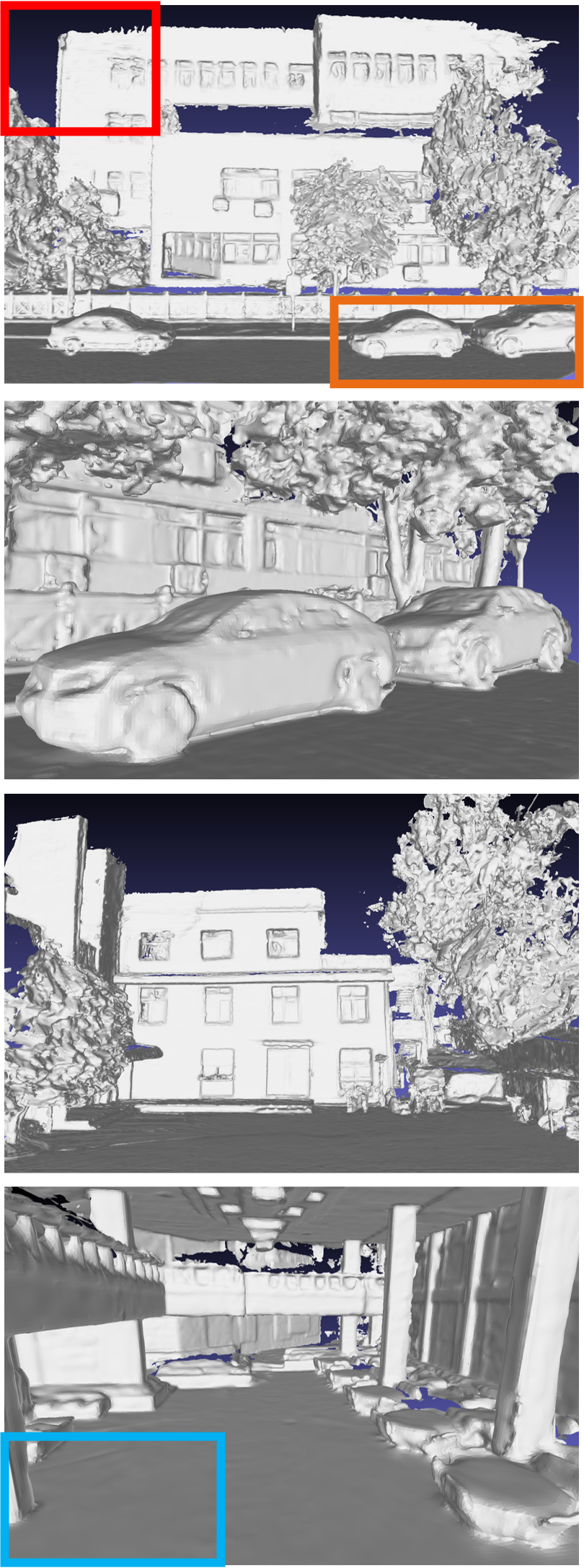} 
\label{fig:Our_mesh}}
\subfloat[2DGS\cite{huang20242d}]{\includegraphics[width=0.14\linewidth]{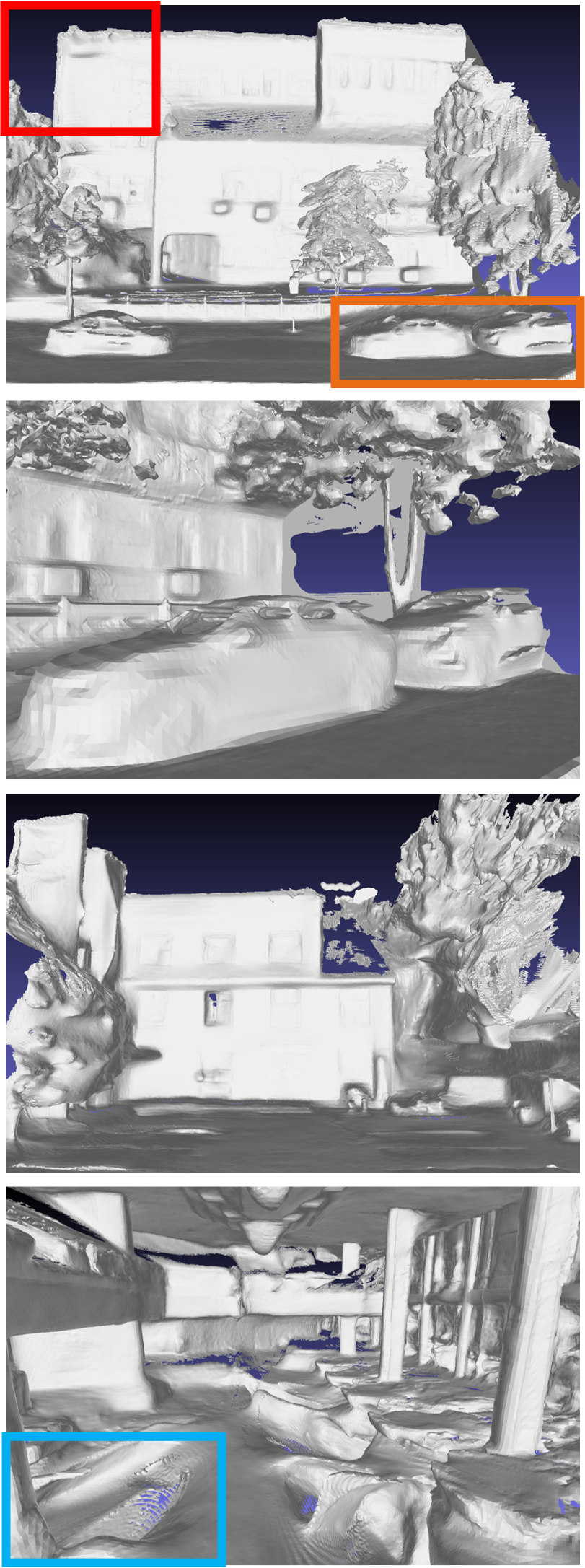} 
\label{fig:2dgs_mesh}}
\subfloat[LIV-GaussMap\cite{hong2024liv}]{\includegraphics[width=0.14\linewidth]{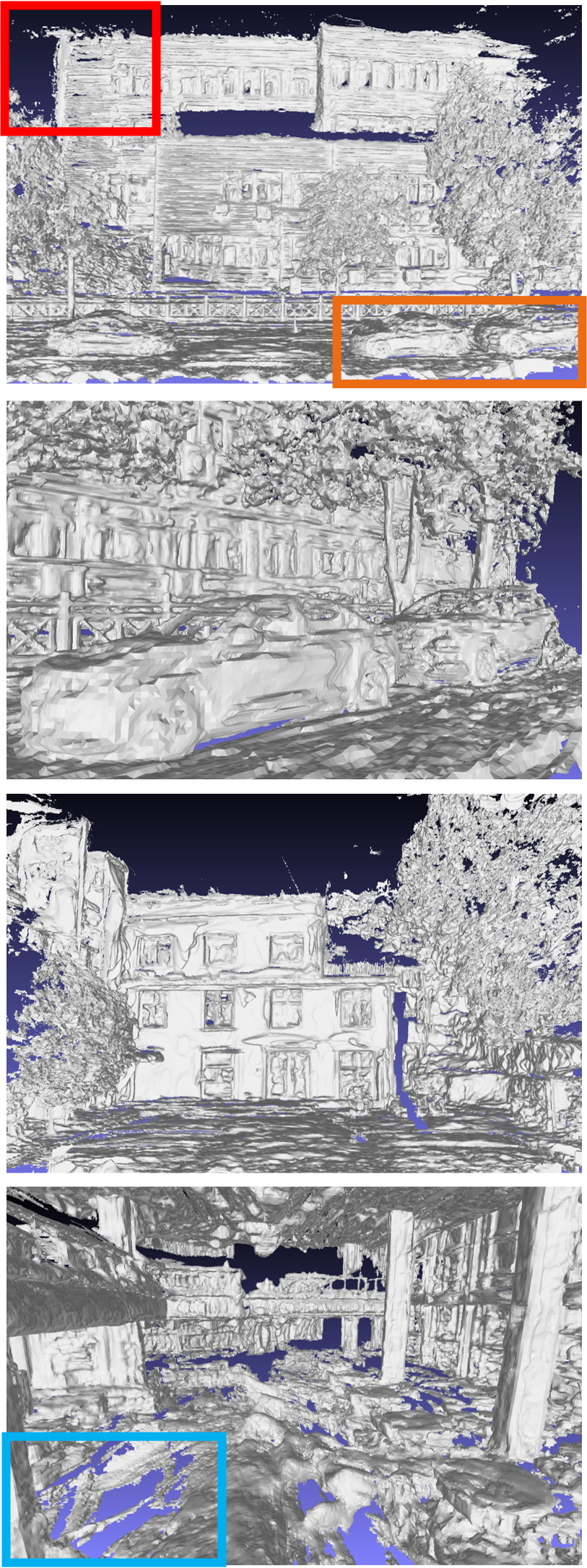} 
\label{fig:liv_mesh}}
\subfloat[GOF\cite{yu2024gaussian}]{\includegraphics[width=0.14\linewidth]{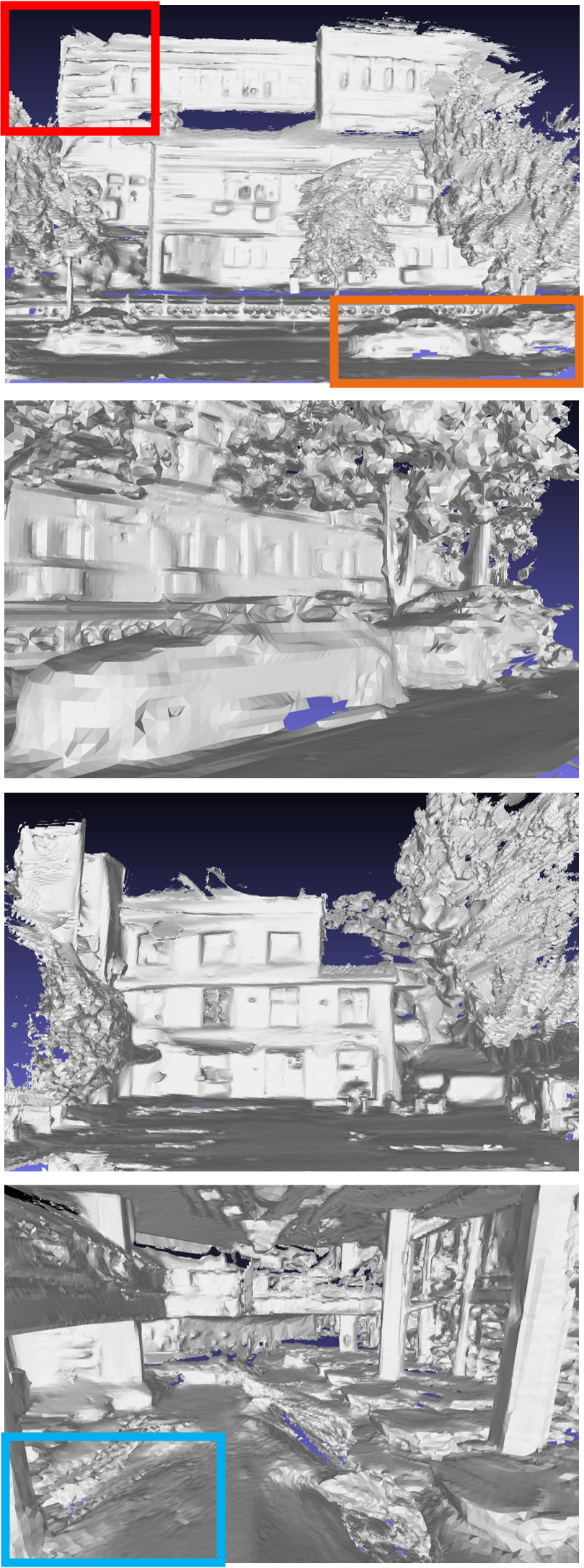} 
\label{fig:GOF_mesh}}
\subfloat[NeRF-LOAM\cite{deng2023nerf}]{\includegraphics[width=0.14\linewidth]{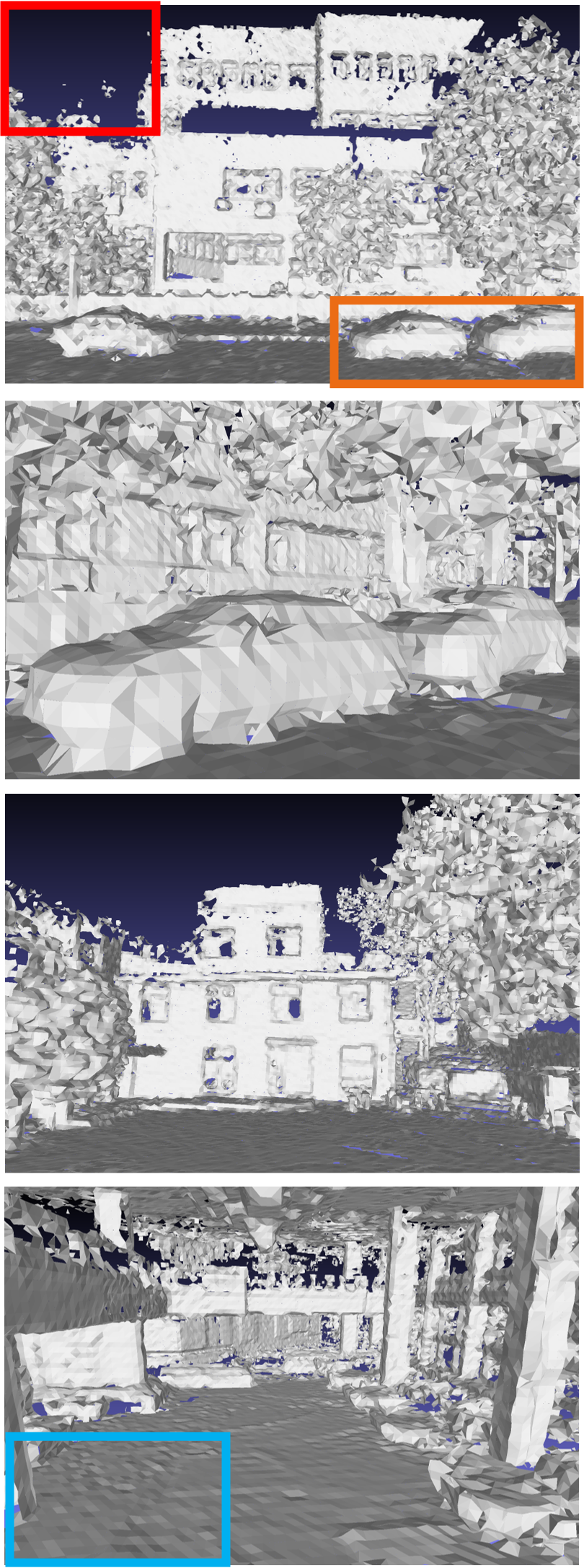} 
\label{fig:nf_mesh}}
\subfloat[VDBFusion\cite{vizzo2022vdbfusion}]{\includegraphics[width=0.14\linewidth]{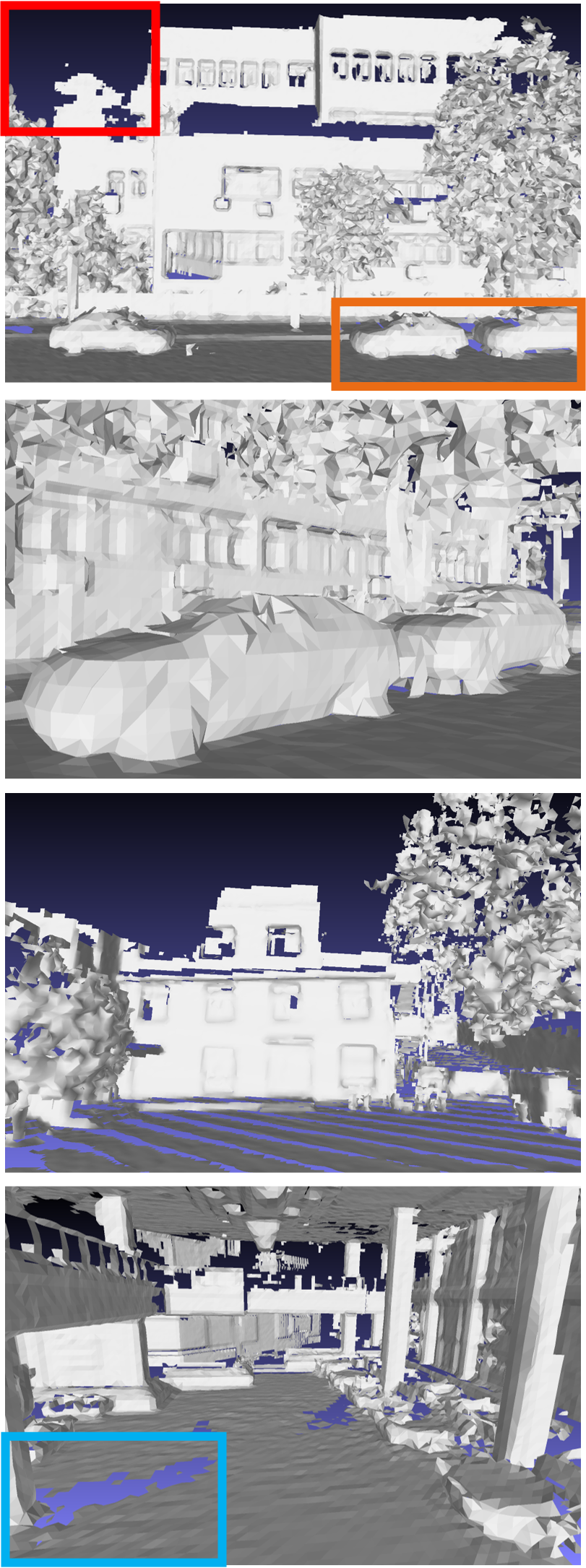} 
\label{fig:vdb_mesh}}
\caption{Geometric quality comparison of different methods in outdoor and indoor scenes. Each row shows the results of a scene using different methods, and the colorized boxes emphasize the surface details.}
\label{fig:geometric_quality}
\end{figure*}

\begin{table}[!t]
	\caption{Dataset Details}
	\centering 
	\resizebox{0.45\textwidth}{!}{
        \begin{threeparttable}
		\begin{tabular}{cccccc}
			\toprule 
			Scene & Description &\# Images &\makecell{\# LiDAR \\ Frames}& \makecell{Trajectory \\ Length (m)} & Area ($\text{m}^\text{2}$) \\
			\midrule 
			1 & Street & 300 & 861 & 59.15 & 449.42 \\
			2 & Street & 568 & 828 & 36.19 & 396.66 \\
			3 & Campus & 393 & 522 & 17.72 & 713.50 \\
			4 & Campus & 454 & 320 & 12.71 & 463.65 \\
			5 & Campus & 2148 & 1640 & 103.30 & 2141.50 \\
			6 & Indoor & 977 & 704 & 33.14 & 959.51 \\ 
			\bottomrule 
		\end{tabular}
        \end{threeparttable}
	}
	\label{table:Dataset}
\end{table}

\begin{table}[!t]
	\caption{Parameter Settings}
	\centering 
	\resizebox{0.45\textwidth}{!}{
        \begin{threeparttable}
		\begin{tabular}{cccccccccc}
			\toprule  
			$\lambda_\text{GMM}$ & $\lambda_\text{d}$ & $\lambda_\text{n}$ & $K$ & $\sigma$ & $\alpha$  & $\tau$ & $\omega_\text{growth}$ & $\omega_\text{scale}$ & $\omega_\text{pruning}$ \\
			\midrule
			1.0 & 0.1 & 0.1 & 4 & 0.1 & 0.5 & 0.01 & 0.4 & 0.0002 & 0.003 \\
			\bottomrule  
		\end{tabular}
        \end{threeparttable}
	}
	\label{table:Parameter}
\end{table}

\section{Experiments}
\subsection{Experimental Setup}
We develop a platform for data collection, as shown in Fig. \ref{fig:first_image}(f). This device consists of two HESAI XT32 LiDAR sensors, an Insta360 ONE RS camera, and an Alubi LPMS-IG1 IMU. We collect six scenes in both outdoor and indoor environments, covering a range of unbounded and sparse-view areas. The details are provided in Tab. \ref{table:Dataset}.
Furthermore, we use a train/test split for the datasets, with every 8th photo reserved for testing. The learning rates are set the same as 3DGS \cite{kerbl20233d}. All experiments are conducted using the parameters provided in Tab. \ref{table:Parameter} and run on a single RTX-4090 GPU.

To evaluate our method quantitatively, We consider both geometric and rendering quality. For geometric quality, we employ metrics such as accuracy, completeness, Chamfer-L1 distance, recall, precision, and F-score. The recall, precision, and F1-score are computed as percentages using a threshold of 20 cm, following the evaluation criteria used in studies \cite{zhong2023shine,deng2023nerf}.  
These metrics are computed by comparing result meshes with the reference precise and dense point clouds obtained by the platform and carefully denoised, consistent with the evaluation methods of studies \cite{vizzo2022vdbfusion,yu2023nf}. For rendering quality, we use the standard metrics such as the Peak Signal-to-Noise Ratio (PSNR) and Structural Similarity Index (SSIM). Additionally, we measure the training time and the number of model components.

\definecolor{first_color}{HTML}{FFA5A5}      % 1st
\definecolor{second_color}{HTML}{FFD2A5}      % 2nd
\definecolor{third_color}{HTML}{FFF9B7}      % 3rd

\begin{table}[!t]
	\caption{Geometric Quality Comparison }
	\centering 
	\resizebox{0.43\textwidth}{!}{
		\begin{threeparttable} 
\begin{tabular}{clcccccc}
\toprule
\multicolumn{2}{c}{Method}&Acc.\textsuperscript{1}$\downarrow$&Comp.$\downarrow$& C-L1$\downarrow$&Re.$\uparrow$&Pre.$\uparrow$&F1$\uparrow$\\
\midrule 
\multirow{4}{*}{\rotatebox{90}{LiDAR}}&SLAMesh & 11.77 & \cellcolor[HTML]{FFF9B7}10.12\textsuperscript{2} & 10.92 & 77.30 & 90.74 & 83.44 \\
&VDBFusion & \cellcolor[HTML]{FFD2A5}7.27 & 13.65 & 10.47 & \cellcolor[HTML]{FFD2A5}88.96 & 89.51 & 88.95 \\ 
&SHINE-Mapping & \cellcolor[HTML]{FFF9B7}8.82 & 10.15 & \cellcolor[HTML]{FFF9B7}9.50 & 85.09 & \cellcolor[HTML]{FFF9B7}94.03 & \cellcolor[HTML]{FFF9B7}89.05 \\ 
&NeRF-LOAM & 9.02 & \cellcolor[HTML]{FFD2A5}7.47 & \cellcolor[HTML]{FFD2A5}8.22 & \cellcolor[HTML]{FFF9B7}85.17 & \cellcolor[HTML]{FFD2A5}95.62 & \cellcolor[HTML]{FFD2A5}90.04 \\
\midrule  
\multirow{10}{*}{\rotatebox{90}{LiDAR+Visual}} &2DGS  & 13.98 & 18.60 & 16.28 & 71.65 & 80.85 & 75.82 \\
&Gaussian Surfels & 14.25 & 35.95 & 25.10 & 70.47 & 48.33 & 57.29 \\
&PGSR & 13.52 & 19.63 & 16.58 & 72.66 & 73.67 & 73.08 \\
&RaDe-GS & 13.28 & 17.13 & 15.22 & 73.42 & 78.59 & 75.88 \\
&LIV-GaussMap  & 14.17 & 15.48 & 14.80 & 72.54 & 80.33 & 76.18 \\
&GOF & 14.27 & 15.37 & 14.83 & 70.55 & 80.20 & 75.04 \\
&SuGaR & 12.88 & 15.62 & 14.23 & 75.02 & 81.86 & 78.23 \\
&Trim2DGS & 14.65 & 33.75 & 25.30 & 68.64 & 61.12 & 64.09 \\
&Trim3DGS & 14.47 & 16.28 & 15.37 & 71.09 & 77.52 & 74.08 \\
&LI-GS (Ours)& \cellcolor[HTML]{FFA5A5}4.37 & \cellcolor[HTML]{FFA5A5}4.63 & \cellcolor[HTML]{FFA5A5}4.51 & \cellcolor[HTML]{FFA5A5}97.49 & \cellcolor[HTML]{FFA5A5}98.65 & \cellcolor[HTML]{FFA5A5}98.07  \\
\bottomrule  
\end{tabular} 
    \begin{tablenotes}
        \footnotesize
        % 1 acc
        \item \textsuperscript{1} Accuracy, completeness, and Chamfer-L1 distance are reported in cm. Recall, precision, and F1-score are calculated as \% using a 20 cm error threshold.
        % 2 color
        \item \textsuperscript{2} Best results are highlighted as \colorbox{first_color}{1st}, \colorbox{second_color}{2nd}, and \colorbox{third_color}{3rd}.
    \end{tablenotes}
		\end{threeparttable}  
	}
	\label{table:GeometricQuality}
\end{table}

\begin{table}[!t]
	\caption{Rendering Quality Comparison }%标题
	\centering
	\resizebox{0.43\textwidth}{!}{
		\begin{threeparttable} 
\begin{tabular}{lccccc}
\toprule 
\multirow{2}{*}{Method} & \multicolumn{2}{c}{Train} & \multicolumn{2}{c}{Test} & \multirow{2}{*}{\makecell{Training \\ Time$\downarrow$}} \\
& PSNR$\uparrow$ & SSIM$\uparrow$ & PSNR$\uparrow$ & SSIM$\uparrow$ & \\
\midrule  
2DGS & \cellcolor[HTML]{FFF9B7}27.35\textsuperscript{*} & 0.888 & 27.18 & 0.884 & \cellcolor[HTML]{FFF9B7}17m \\ 
Gaussian Surfels & 26.41 & 0.856 & 26.34 & 0.854 & \cellcolor[HTML]{FFD2A5}12m40s \\ 
LIV-GaussMap & 27.29 & \cellcolor[HTML]{FFD2A5}0.897 & \cellcolor[HTML]{FFD2A5}28.29 & \cellcolor[HTML]{FFD2A5}0.899 & \cellcolor[HTML]{FFA5A5}11m12s \\ 
GOF & \cellcolor[HTML]{FFA5A5}28.57 & \cellcolor[HTML]{FFA5A5}0.905 & \cellcolor[HTML]{FFA5A5}28.39 & \cellcolor[HTML]{FFA5A5}0.901 & 60m46s \\
SuGaR & 26.73 & 0.882 & 27.31 & 0.883 & 59m24s \\
Trim2DGS & 25.33 & 0.872 & 25.24  & 0.87 & 32m1s \\
LI-GS (Ours)& \cellcolor[HTML]{FFD2A5}27.80 & \cellcolor[HTML]{FFF9B7}0.889 & \cellcolor[HTML]{FFF9B7}27.60 & \cellcolor[HTML]{FFF9B7}0.885 & 33m16s\\
\bottomrule 
\end{tabular}
    \begin{tablenotes}
        \footnotesize
        \item \textsuperscript{*} Best results are highlighted as \colorbox{first_color}{1st}, \colorbox{second_color}{2nd}, and \colorbox{third_color}{3rd}.
    \end{tablenotes}
			
		\end{threeparttable} 
	}
	\label{table:NVSQuality}
\end{table}

\subsection{Comparative Study}
We compare our approach with four mapping methods that utilize only LiDAR (SLAMesh \cite{ruan2023slamesh}, VDBFusion \cite{vizzo2022vdbfusion}, SHINE-Mapping \cite{zhong2023shine}, and NeRF-LOAM \cite{deng2023nerf}), as well as nine state-of-the-art Gaussian-based surface reconstruction methods.
To ensure the correct scaling of result meshes and enhance the mapping capability of Gaussian-based methods in large-scale scenes for fair comparison, we initialize these methods using colorized point clouds, following the ideas described
in studies \cite{hong2024liv, tao2024silvr, cui2024letsgo}. Additionally, we also incorporate sky masks in these methods.

As shown in Tab. \ref{table:GeometricQuality}, our approach outperforms all baseline methods across all metrics. 
The performance of our approach is further shown in Fig. \ref{fig:geometric_quality}. When comparing Fig. \ref{fig:geometric_quality}(a) with (e) and (f), it is evident that Gaussian-based methods can reconstruct more comprehensive surfaces, and this can be attributed to the using of visual information. Furthermore, when comparing Fig. \ref{fig:geometric_quality}(a) with (c) and (d), it is observed that the introduction of LiDAR normalization mitigates the excessive impact of the photometric loss on the positions and shapes of Gaussians, thereby enhancing reconstruction quality. Additionally, when comparing Fig. \ref{fig:geometric_quality}(a) with (b), it is evident that the normal-depth consistency regularization of 2DGS results in a loss of mesh details in large-scale scenes.

\begin{figure}[!t]
\centering
\subfloat[Rendered RGB]{\includegraphics[width=0.25\linewidth]{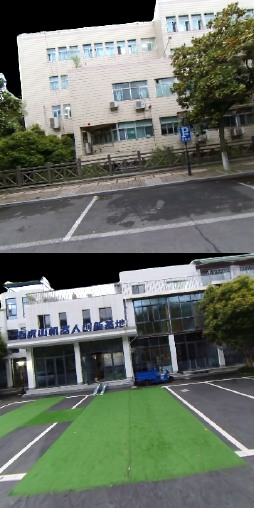} 
\label{fig:nvs}}
\subfloat[w/ GMM loss]{\includegraphics[width=0.25\linewidth]{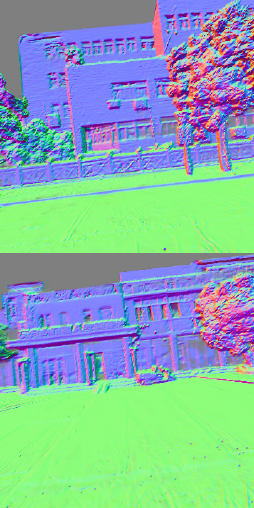} 
\label{fig:loss_compare_ours}}
\subfloat[w/o GMM loss]{\includegraphics[width=0.25\linewidth]{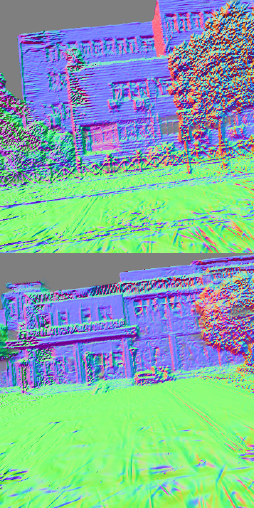} 
\label{fig:loss_compare_wo_gmm_loss}}
\caption{Results of our method in two scenes. (a) Rendered RGB images. (b)-(c) Comparison of rendered normal images with and without GMM loss.}
\label{fig:loss_compare}
\end{figure}

As shown in Tab. \ref{table:NVSQuality}, our method achieves comparable results with the current Gaussian-based methods in terms of rendering quality, as shown in Fig. \ref{fig:loss_compare}(a). Our method demonstrates improved rendering quality compared to 2DGS, which can be attributed to the integration of accurate LiDAR normalization.
As for training time, our method is slightly slower than 2DGS due to the requirement of nearest neighbor search in our GMM normalization.

\begin{table}[!t]
	\caption{Ablation Study on Initialization and Normalization}%标题
	\centering
	\resizebox{0.43\textwidth}{!}{
		\begin{threeparttable} 
\begin{tabular}{lccccc}
\toprule 
\makecell[l]{Initialization \\ Method}& \makecell{Normalization \\ with GMM Loss} & Acc.\textsuperscript{*}$\downarrow$ &Comp.$\downarrow$& C-L1$\downarrow$&F1$\uparrow$\\ 
\midrule 
SfM & \checkmark & 5.18 & 12.68 & 8.93 &  93.15 \\ 
Colorized PC & \checkmark & 4.95 & 8.96 & 6.98 & 95.00 \\
GMM & \texttimes & 7.10 & 15.08 & 11.10 & 89.35 \\
GMM & \checkmark & \textbf{4.37} & \textbf{4.63} & \textbf{4.51} & \textbf{98.07} \\
\bottomrule 
\end{tabular}
\begin{tablenotes}
    \footnotesize
    \item \textsuperscript{*} Accuracy, completeness, and Chamfer-L1 distance are reported in cm. F1-score is calculated as \% using a 20 cm error threshold.
\end{tablenotes}
			
		\end{threeparttable} 
	}
	\label{table:AblationGeo}
\end{table}

\subsection{Ablation Study}

\subsubsection{Initialization}
\begin{figure}[!t]
    \centering
    \includegraphics[width=0.86\linewidth]{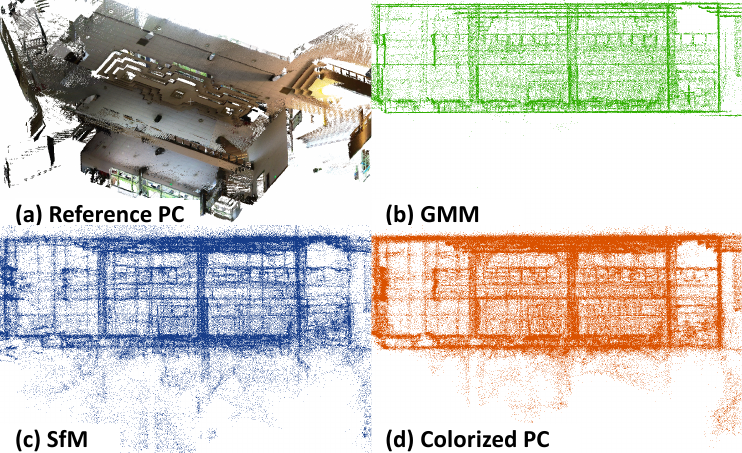}
    \caption{Comparison of initialization methods in the indoor scene. (a) The reference colorized point cloud. (b)-(d) The positions of Gaussians of three initialization methods. }
    \label{fig:init_compare}
\end{figure}

We compare three initialization methods: SfM points generated by Colmap-PCD \cite{bai2024colmap} (SfM), colorized LiDAR point clouds (Colorized PC), and our GMM-based initialization method (GMM). 
Tab. \ref{table:AblationGeo} shows that GMM-based initialization method achieves better geometric quality. 
Fig. \ref{fig:init_compare} presents a clear comparison of the initialization methods. The indoor scene exhibits the issue of ground reflection, as shown in Fig. \ref{fig:first_image}(g). It is evident that inappropriate initialization methods can result in incorrect convergence. Specifically, the Gaussians on the ground become transparent, while numerous Gaussians appear underground (Fig. \ref{fig:init_compare}(c) and (d)). 
Similar issues occur in other Gaussian-based reconstruction methods, as depicted in the last row of Fig. \ref{fig:geometric_quality}, resulting in ground depressions in the result mesh. In contrast, GMMs provide more accurate initial models, leading to a correct convergence. Fig. \ref{fig:init_compare}(b) illustrates that our method does not have Gaussians underground. As shown in Fig. \ref{fig:geometric_quality}(a), our method produces a smooth and complete ground mesh. Moreover, as shown in Tab. \ref{table:Iter1_nvs}, the integration of LiDAR can enhance the initialization for novel view synthesis.

\begin{table}[!t]
	\caption{Iter1 Rendering Quality Using Different Initialization Methods}%标题
	\centering
 \resizebox{0.3\textwidth}{!}{
		\begin{threeparttable} 
                \begin{tabular}{lccc}
                \toprule 
                Initialization Method & L1$\downarrow$ &PSNR$\uparrow$ & SSIM$\uparrow$\\ 
                \midrule 
                SfM & 0.207 & 11.74 & 0.139 \\ 
                Colorized PC & 0.069 & 17.14 & 0.616 \\
                GMM & \textbf{0.065} & \textbf{17.21} & \textbf{0.618} \\
                \bottomrule 
                \end{tabular}
		\end{threeparttable} 
	 }
	\label{table:Iter1_nvs}
\end{table}

\begin{figure}[!t]
    \centering
    \includegraphics[width=0.86\linewidth]{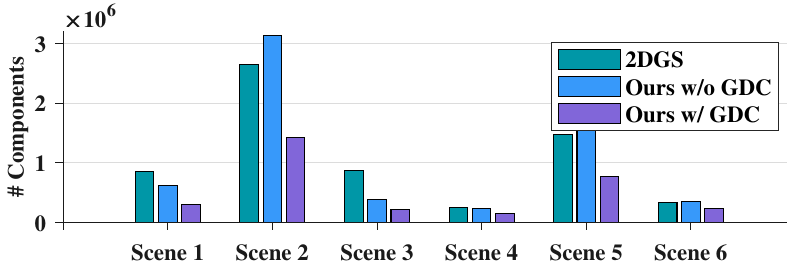}
    \caption{Comparison of the numbers of model components in six scenes.}
    \label{fig:ablation_components}
\end{figure}

\begin{figure}[!t]
    \centering
    \includegraphics[width=0.88\linewidth]{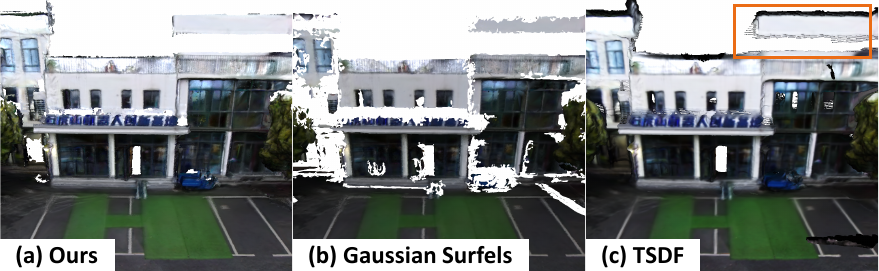}
    \caption{Comparison of three different mesh extraction methods.}
    \label{fig:ablation_meshing}
\end{figure}

\begin{table}[!t]
	\caption{Ablation Study On Mesh Extraction}
	\centering
	\resizebox{0.38\textwidth}{!}{
		\begin{threeparttable} 
\begin{tabular}{lcccc}
\toprule
\makecell[l]{Mesh Extraction\\Method} & Acc.\textsuperscript{*}$\downarrow$ &Comp.$\downarrow$& C-L1$\downarrow$&F1$\uparrow$\\ 
\midrule 
TSDF & 12.48 & 11.90 & 12.23 & 83.02 \\ 
Gaussian Surfels & \textbf{3.42} & 17.35 & 10.37 & 90.11 \\
Ours & 4.37 & \textbf{4.63} & \textbf{4.51} & \textbf{98.07} \\
\bottomrule 
\end{tabular}
\begin{tablenotes}
    \footnotesize
    \item \textsuperscript{*} Accuracy, completeness, and Chamfer-L1 distance are reported in cm. F1-score is calculated as \% using a 20 cm error threshold.
\end{tablenotes}
			
		\end{threeparttable} %添加此处
	}
	\label{table:Mesh_extraction}
\end{table}

\subsubsection{Optimization}
Our method is evaluated both with and without GMM normalization, and the geometric quality is presented in Tab. \ref{table:AblationGeo}. The results indicate that our method achieves better geometric quality when GMM normalization is applied. Fig. \ref{fig:gmm_gs} and Fig. \ref{fig:loss_compare} show the performance of GMM normalization in reducing noise and producing smooth surfaces.
Furthermore, we evaluate our method both with and without Geometry-aware Density Control (GDC) and visualize the number of components in each scene in Fig. \ref{fig:ablation_components}. On average, the components are reduced by $45.22\%$, showing the effective removal of redundant and inaccurate Gaussians by GDC.

\subsubsection{Mesh Extraction}
We compare three mesh extraction methods: TSDF from 2DGS, the cutting and meshing method from Gaussian Surfels \cite{dai2024high}, and our coarse-to-fine method. Although the method from Gaussian Surfels achieves slightly better accuracy, it removes too many sampled points, resulting in an incomplete mesh (Fig. \ref{fig:ablation_meshing}(b)) and inferior completeness (Tab. \ref{table:Mesh_extraction}). Additionally, TSDF generates incorrect sampled points near depth discontinuities, as indicated by the colorized box in Fig. \ref{fig:ablation_meshing}(c). Our mesh achieves a good balance between accuracy and completeness.

\section{Conclusion}
This paper introduces LI-GS, a reconstruction system that incorporates LiDAR with Gaussian Splatting to improve geometric accuracy in large-scale scenes. The precise LiDAR measurements are leveraged in various essential steps of the system, including initialization, regularization, density control, and mesh extraction. Additionally, we propose an incremental multimodal modeling approach with plane constraints for generating GMMs. Experiments validate the outstanding performance of LI-GS in 3D reconstruction. Our future work will involve the application of our method in Simultaneous Localization and Mapping (SLAM).

\bibliographystyle{IEEEtran}
\bibliography{ref}

\vfill

\end{document}